# Segmentation Framework for Heat Loss Identification in Thermal Images: Empowering Scottish Retrofitting and Thermographic Survey Companies

Md Junayed Hasan[1][0000-0003-4578-952X], Eyad Elyan[2][0000-0002-8342-9026], Yijun Yan[1][0000-0003-0224-0078], Jinchang Ren[1] [0000-0001-6116-3194], Md Mostafa Kamal Sarker[3][0000-0002-4793-6661]

[1] National Subsea Centre, Robert Gordon University, Aberdeen, AB21 0BH, UK
[2] School of Computing, Robert Gordon University, Aberdeen, AB10 7AQ, UK
[3] Institute of Biomedical Engineering, University of Oxford, Oxford, OX3 7DQ, UK

**Abstract.** Retrofitting and thermographic survey (TS) companies in Scotland collaborate with social housing providers to tackle fuel poverty. They employ ground-level infrared (IR) camera-based-TSs (GIRTSs) for collecting thermal images to identify the heat loss sources resulting from poor insulation. However, this identification process is labor-intensive and time-consuming, necessitating extensive data processing. To automate this, an AI-driven approach is necessary. Therefore, this study proposes a deep learning (DL)-based segmentation framework using the Mask Region Proposal Convolutional Neural Network (Mask RCNN) to validate its applicability to these thermal images. The objective of the framework is to automatically identify, and crop heat loss sources caused by weak insulation, while also eliminating obstructive objects present in those images. By doing so, it minimizes labor-intensive tasks and provides an automated, consistent, and reliable solution. To validate the proposed framework, approximately 2500 thermal images were collected in collaboration with industrial TS partner. Then, 1800 representative images were carefully selected with the assistance of experts and annotated to highlight the target objects (TO) to form the final dataset. Subsequently, a transfer learning strategy was employed to train the dataset, progressively augmenting the training data volume and fine-tuning the pre-trained baseline Mask RCNN. As a result, the final fine-tuned model achieved a mean average precision (mAP) score of 77.2% for segmenting the TO, demonstrating the significant potential of proposed framework in accurately quantifying energy loss in Scottish homes.

## 1   Introduction

Fuel poverty is predicted to affect about 39% of Scottish households after a significant increase in energy prices was announced in April 2023. Therefore, the Scottish government has made it a top priority to reduce carbon emissions from homes and businesses to mitigate the fuel poverty. The Scottish Fuel Poverty Act (SFPA) aims to increase energy efficiency and reduce carbon footprint in all infrastructures by 2040 [1]. Additionally, it is imperative to prioritize the energy efficiency of buildings, as it aids in securing funding for emerging technologies like heat pumps, insulation, and retrofitting. Specialized companies that focus on retrofitting and conducting thermographic surveys (TSs) collaborate with social housing providers to ensure compliance with the Energy Efficiency Standard for Social Housing (EESSH) policy [2] . They detect buildings that require retrofitting by using thermal images gathered by ground-level infrared (IR) camera-based-TSs (GIRTSs) [3]. However, the current challenge is the need to manually analyze those images after collection to identify sources of heat loss and eliminate obstructive objects. This process is labor-intensive, time-consuming, and relies heavily on domain experts. As a result, this hinders scalability and limits the utilization of cloud-based statistical thermal profile analyzer toolkits. Therefore, this research aims to develop a deep learning (DL) based-automated solutions to identify the target objects (TO) related to the actual heat loss in GIRTS-thermal images, streamlining data analysis and reducing the mentioned labor-intensive tasks.

Researchers have addressed similar problem types by utilizing vehicle and drone-mounted IR cameras to capture thermal images of building facades and applying various machine learning (ML) algorithms to detect thermal bridges. Macher et al. [4] used a vehicle-mounted camera to create a thermographic 3D point cloud and successfully detect thermal bridges under balconies and between levels. However, this method has limitations such as difficulty in extracting ground-level and basement windows and the inability to detect windows hidden by foliage or other objects. Using drones for thermographic assessments has become more popular as they allow for the entire outside of a building to be captured and there is less interference caused by obstructions. Rakha et al. [5] proposed a thermal drone-based system to locate thermal anomalies in building envelopes and claimed 75% precision, while Mirzabeigi et al. [6] developed a computer vision algorithm and drone flight path to detect thermal anomalies but lack quantitative data on the effectiveness. Kim et al. [7]used neural networks to identify thermal bridges in terrestrial thermographic images with an average precision of 89% and recall of 87%. This study examines the difficulties encountered when using thresholding and histogram methods (as discussed above) to identify TO in non-stationary thermography research in panorama settings. However, these studies face two main challenges. Firstly, the use of aerial thermographic survey (ATS) and vehicle-based moving thermographic survey (VMTS) makes it difficult to capture suitable vantage points for identifying thermal bridges and sources of heat loss. These approaches cover multiple infrastructures and encounter varying conditions, such as weather, lighting, and pose fluctuations, making it challenging to accurately identify the sources of heat loss amid obstructive objects. Secondly, to address the complexity resulting from the first challenge, existing research has employed numerous manual thresholding techniques to remove obstructive objects from ATS/VMTS-thermal images and identify sources of heat loss. However, due to the varying conditions present in these thermal images, the accuracy of thresholding is not optimal. Consequently, ML solutions developed to identify the sources of heat loss struggle to perform effectively and robustly.

To tackle the challenges associated with vantage points in ATS/VMTS-based-thermal image analysis, we conducted the GIRTS in partnership with an industrial TS partner. As a result, we obtained a dataset of 2500 thermal images, from which we carefully selected 1800 images with the help of domain experts and annotated the target objects (TO) to create a customized dataset called GIRTSD. Then, to automate the identification, detection, and removal of potential sources of heat loss as well as obstructive objects, we employed the Mask Region Proposal Convolutional Neural Network (Mask RCNN [8]), a deep learning (DL)-based segmentation algorithm, eliminating the need for manual thresholding and enhancing accuracy. However, since the GIRTSD might not be sufficient for training due to variations in object shapes, forms, and weather conditions, we employed transfer learning strategies. Specifically, we fine-tuned a pre-trained Mask RCNN model from the Microsoft Common Object in Context (MSCOCO) dataset [9], which includes 80 object categories. After preparing the fine-tuned model using a limited set of samples from our GIRTSD, we progressively expanded the training data, integrated diverse image augmentation techniques, and employed transfer learning strategies to improve the model's performance [10, 11]. Thus, after conducting extensive ablation studies, we selected the optimal fine-tuned model, which achieved an impressive mean average precision (mAP) of 77.2% for segmenting the TO. The main contributions of this study are highlighted below.

1. We introduced the ground-level infrared camera-based thermographic survey (GIRTS) for collecting thermal images, resulting in a custom dataset named GIRTSD. This dataset comprises 1800 annotated thermal images representing 7 distinct target objects (TO), including possible sources of heat loss and obstructive objects. Unlike aerial/vehicle-based moving thermographic survey, GIRTS accurately captures various TO regardless of object shapes, forms, and weather conditions.
2. To overcome the limitations of existing research in thermal image-based heat loss source detection, specifically manual thresholding, we propose a Mask RCNN-based framework. This framework detects and crops potential heat loss sources while eliminating obstructive objects, resulting in reduced manual labor and improved efficiency in identifying energy losses in buildings.

The rest of the paper is organized as follows: Section 2 refers to the experimental set up and methodology, Section 3 discusses the experimental finding, and finally, Section 4 concludes the paper.

## 2    Experimental setup and framework strategy

Fig. 1 depicts the complete framework for the validation of our case study. The process begins with data collection by GIRTSs, followed by dataset creation with the help of our industrial TS partner. The GIRTSD creation is the initial step, followed by offline training-testing to refine the proposed Mask RCNN based segmentation framework. Once satisfactory performance is achieved, the optimal model is selected for online evaluation.

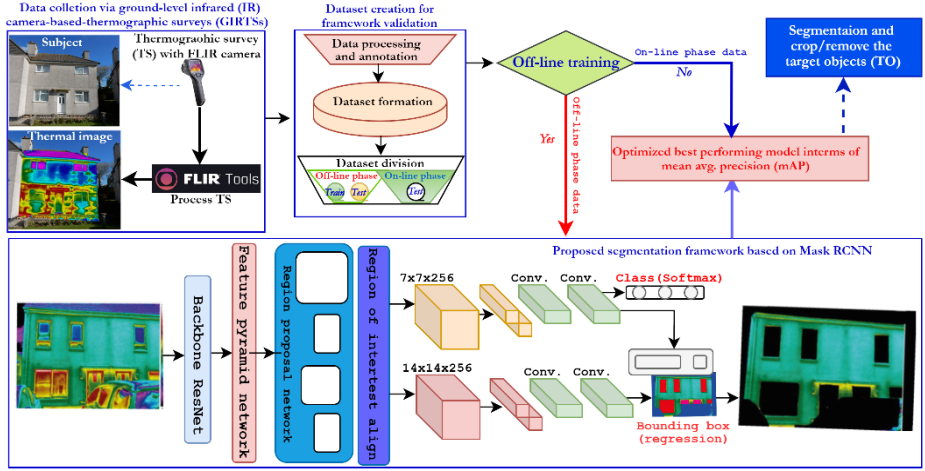

**Fig. 1**. Proposed segmentation framework for heat loss identification in thermal images.

### 2.1 Data collection and dataset creation

**Table 1.** Details of our custom thermal image dataset, GIRTSD

| | | Off-line phase | | On-line phase |
|---|---|---|---|---|
| | | Train | Test | Evaluation set |
| TO | | Number of instances/ TO | | |
| **Heat loss sources** | Window | 807 | 89 | 99 |
| | Door | 278 | 31 | 34 |
| **Obstructive objects** | Fence | 61 | 7 | 8 |
| | Tree | 85 | 9 | 10 |
| | Bin | 116 | 13 | 14 |
| | Road | 111 | 12 | 13 |
| | Other | 851 | 94 | 105 |

Total 1800 thermal images, with 7 target objects (TO)

TSs are non-invasive methods used to identify insulation issues and heat loss in buildings [12]. These surveys utilize thermal images to detect air leakage, moisture infiltration, and structural defects, improving energy efficiency and reducing environmental impact. Trained professionals with expertise in both technology and building science are required for accurate surveys. In our study, we collaborated with an industrial TS partner in Scotland and collected thermal images using GIRTSs and FLIR E60bx cameras [13]. The high-resolution infrared detectors and multi-spectral dynamic imaging feature of these cameras provide detailed thermal images with a temperature range of −20℃ to +350℃. FLIR Tools software is used to process, and analyse these surveys, and export those as the thermal images. Then, with the help of our TS partner, we selected 1800 diverse images ranging from 320 × 256 to 1920 × 1536 pixels. The subsequent step involved annotating those images to highlight heat loss sources and obstructive objects. The annotation process is time-consuming but crucial for instance segmentation. We used LabelMe as the annotation tool, following the COCO dataset format, and obtained segmented ground truth masks in JSON format [9]. Each image was resized to $512 \times 512$ for training, maintaining a unified aspect ratio. Further details about our dataset – GIRTSD can be found in Table 1.

### 2.2 Model training strategy, evaluation, and optimization

To identify the best segmentation framework based on the COCO evaluation metric (mAP [9]), we conducted a two-phase process. Firstly, we performed off-line phase training to determine the top-performing framework. Then, we evaluated its performance using the online phase-evaluation data. The reported performance indicators in Table 2 are calculated during the off-line phase training and test data.

Given the limited number of samples, constructing a generalized DL model posed challenges. As a solution, we adopted the Mask RCNN model, denoted as Mb, which had been pretrained on the 80-class MS-COCO dataset, as our framework's baseline. While preserving the initial layers that capture low-level features, we retrained the remaining components responsible for high-level features, effectively producing a modified version of Mb. To customize Mb, we leveraged TensorFlow, Keras, OpenCV, and Python 3, utilizing the open-source MMDetection toolbox [14].

**Table 2.** Ablation study during off-line phase.

| Models | Backbone - base model | Data Aug. | Training data vol. | $mAP^{50\text{-}95}$ Train | $mAP^{50\text{-}95}$ Test |
|---|---|---|---|---|---|
| M1/M2 | R50-Mb | No/Yes | 20% | 89.2/91.1 | 59.2/60.1 |
| M3/M4 | R101-Mb | No/Yes | 20% | 92.2/94.1 | 60.9/61.4 |
| M5/M6 | R101-M4 | No/Yes | 40% | 94.5/95.1 | 64.3/65.9 |
| M7/M8 | R101-M4 | No/Yes | 60% | 95.7/96.3 | 67.9/69.2 |
| M9/M10 | R101-M4 | No/Yes | 80% | 96.8/97.1 | 70.7/72.9 |
| M11/**M12** | R101-M4 | No/**Yes** | **100%** | 97.5/**98.2** | 75.8/**78.7** |

For the classification tasks within the segmentation pipeline, we employed ResNet-50 (R50) and ResNet-101(R101) as backbone models, using stochastic gradient descent (SGD) [15] and Adam optimizers [16]. This resulted in the generation of four models (M1 - M4) to explore the retraining of Mb on our custom dataset while considering the inclusion or exclusion of image augmentation techniques. The objective was to determine the optimized framework [17]. During this phase, we only utilized 20% of the training data from the offline phase. Once the best-performing model was identified, it was subjected to additional ablation studies and fine-tuning.

In our designed ablation experiments, we aimed to demonstrate the impact of increasing training data on the performance of our model. We initially used only 20% of the data from Table 1 for training, gradually increasing the training data volume by 20% in subsequent experiments. By observing the performance of our model on the off-line phase training and test data, we assessed whether increased data volume resulted in improved model performance.

We evaluated different models (M5 - M12) by considering both with and without image augmentation techniques, using *mAP*. The mask head training region of interest (RoI) was set to 200, while the maximum number of ground truth instances per image was raised from 100 to 512. Furthermore, the incremental training RoI parameter was set to 512. Detailed information regarding our various ablation studies can be found in Table 2.

## 3   Result analysis and discussion

Table 2 presents the *mAP$^{50-95}$* results for both off-line phase training and testing. The *mAP$^{50-95}$* is calculated by averaging accuracy over the intersection over union (IoU) range from 0.5 to 0.95, with a step size of 0.05. Among the models M1-M4, it was observed that M4, with the R101-Mb combination as the backbone-base model along with data augmentation, performed the best. Subsequently, four pairs of models were gradually built from M4, increasing the training data volume to observe performance improvements during the offline training phase. In all experiments, data augmentation consistently led to better performance. Additionally, as the training data volume was incrementally increased by 20%, a gradual improvement in performance was observed. These findings clearly highlight the significance of data diversity and the quantity of training samples in building a generalized and robust model. It is worth noting that as the trained model improved, the performance on the offline test data also improved.

In Table 1, it can be observed that the number of instances/TO varies significantly, which impacts the overall model performance. To validate this observation, a closer examination was conducted on the best performing model from Table 2 (M12), specifically analyzing its off-line phase test *mAP$^{50-95}$* /TO. Detailed analysis can be found in Table 3, and Fig. 2.

**Table 3.** Off-line phase test performance of M12/TO.

| *mAP$^{50-95}$* | *BBOX* | *SEGM* |
|---|---|---|
| Window | 78.9 | 80.3 |
| Door | 74.8 | 77.5 |
| Fence | 59.1 | 62.3 |
| Tree | 57.2 | 47.3 |
| Bin | 71.2 | 72.4 |
| Road | 59.0 | 60.8 |
| Other | 66.7 | 65.3 |

Based on Table 3, it is evident that windows and doors achieved the best results in terms of bounding box detection (BBOX) and segmentation (SEG). These two object categories have a relatively higher number of training instances. On the other hand, the category labeled as 'other', which includes various miscellaneous objects like benches, chairs, flower gardens, and staircases, had the highest number of instances but did not perform well. This can be attributed to the significant variation in pose and shape among these objects, making it challenging for the model to achieve consistent performance. Therefore, along with a larger number of instances, maintaining a unified object shape and category is crucial. For instance, the 'bin' category, despite having only 116 instances, performed impressively due to the relatively consistent shape of the objects within that category.

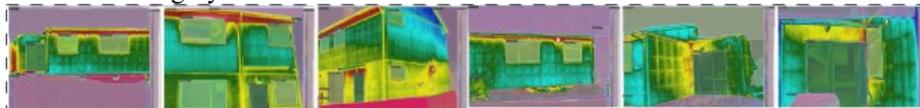

**Fig. 2.** Snapshot of segmentation results.

Having reached this stage, we have chosen the best-performing model, M12, from our offline phase. To assess the overall performance of our framework, we utilized M12 for TO classification and segmentation first. Then, the unwanted objects will be removed based on the segmentation results. During this online phase with the test data, M12 achieved a ***mAP$^{50-95}$*** of 77.2%, which is remarkably close to our performance in the offline phase. Alongside the development of our prototype, we have designed a user-friendly web interface for end-users to evaluate the functionality. This interface allows users to upload GIRTS-thermal images and visually inspect the detected/segmented regions for possible heat loss sources or obstructive objects. It also displays the cropped and cleaned versions of the images, enabling users to save the thermal information from the processed images. Fig. 3 illustrates the activities and workflow of our proposed solution within the web user interface.

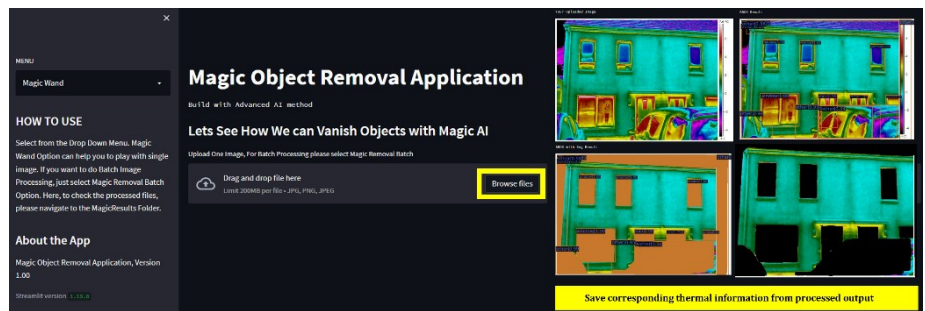

**Fig. 3.** Snapshot of our developed web-interface.

## 4   Conclusions

In this research, we propose an AI-driven solution for ground level infrared camera-based thermographic surveys using thermal images. We implement a Mask RCNN-based deep learning framework to identify, segment, and remove heat sources and unwanted objects. As a result, we created a new thermographic dataset. To overcome the issue of limited thermal images for training, we employed transfer learning strategies and various image augmentation techniques. Through fine-tuning experiments and parameter adjustments, our final model achieved a mAP score of 77.2%. In future research, we aim to improve the training dataset by utilizing generative deep architectures. We also plan to address the challenge of dense clusters of distracting objects that can potentially lead to the model overlooking certain segments. Furthermore, we aim to demonstrate the practical implementation of this solution by developing an open-source tool specifically designed for industries involved in thermographic surveys. Additionally, we have plans to publicly release the GIRTS dataset, enabling further exploration of the potential outcomes achieved through this dataset.

## Acknowledgement

We are thankful to IRT surveys for the thermograph survey support and Data Lab for the funding of this project.